\documentclass{article}

\usepackage{microtype}
\usepackage{graphicx}
\usepackage{multirow}   

\usepackage{subcaption}
\usepackage{booktabs} 

\usepackage{hyperref}
\usepackage{pifont}

\usepackage{tabularx}
\newcolumntype{C}{>{\centering\arraybackslash}X}

\usepackage{icml2026}

\usepackage{amsmath}
\usepackage{amssymb}
\usepackage{mathtools}
\usepackage{amsthm}

\usepackage[capitalize,noabbrev]{cleveref}

\theoremstyle{plain}
\newtheorem{theorem}{Theorem}[section]

\theoremstyle{definition}

\theoremstyle{remark}

\usepackage[textsize=tiny]{todonotes}

\icmltitlerunning{MASA: Rethinking the Representational Bottleneck in LoRA with Multi-A Shared Adaptation}

\begin{document}

\twocolumn[
  \icmltitle{MASA: Rethinking the Representational Bottleneck in LoRA with \\Multi-A Shared Adaptation}



  \icmlsetsymbol{equal}{*}

  \begin{icmlauthorlist}
    \icmlauthor{Qing Dong}{equal,yyy}
    \icmlauthor{Yuntian Tang}{equal,yyy}
    \icmlauthor{Heming Jia}{xxx}
    \icmlauthor{Yunhang Shen}{zzz}
    \icmlauthor{Bohan Jia}{yyy}
    \icmlauthor{Wenxuan Huang}{yyy}
    \icmlauthor{Lianyue Zhang}{yyy}
    \icmlauthor{Jiao Xie}{yyy}
    \icmlauthor{Shaohui Lin}{yyy,zzz}
    \icmlauthor{Rongrong Ji}{zzz}
    
  \end{icmlauthorlist}

  \icmlaffiliation{yyy}{East China Normal University}
  \icmlaffiliation{xxx}{Sanming University}
  \icmlaffiliation{zzz}{Key Laboratory of Multimedia Trusted Perception and Efficient Computing of Ministry of Education of China, Xiamen University, China}

  \icmlcorrespondingauthor{Shaohui Lin}{shaohuilin007@gmail.com}


  \vskip 0.3in
]


\printAffiliationsAndNotice{\icmlEqualContribution}

\begin{abstract}
Low-Rank Adaptation (LoRA) has emerged as a dominant method in Parameter-Efficient Fine-Tuning (PEFT) for large language models, which augments the transformer layer with one down-projection $A$ and one up-projection $B$. However, LoRA's reliance on a single down-projection matrix ($A$) creates a representational bottleneck, as this solitary feature extractor is inherently insufficient for capturing the diverse signals required by complex tasks. This motivates our architectural shift to focus on enriching the feature adaptation to improve the downstream task adaptation ability. We propose \textbf{MASA} (\textbf{M}ulti-\textbf{$A$} \textbf{S}hared \textbf{A}daptation), an architecture that implements a ``multi-$A$, single-$B$'' structure where the multi-$A$ expert ensemble is asymmetrically shared across layers to ensure parameter efficiency. In MASA, these specialized experts capture diverse features, which are then integrated by a single, layer-specific $B$-matrix. The effectiveness and versatility of our method are validated through a comprehensive suite of experiments spanning multi-domain generalization, single-domain specialization, and multi-task reasoning. For example, on the MMLU benchmark, MASA achieves an average accuracy of 59.62\%, outperforming the standard LoRA by 1.08 points (a relative improvement of 1.84\%) while training only 0.52\% of the parameters.
\end{abstract}

\section{Introduction}
\begin{figure*}[t]
\vskip 0.2in
  \centering
  \includegraphics[width=\linewidth]{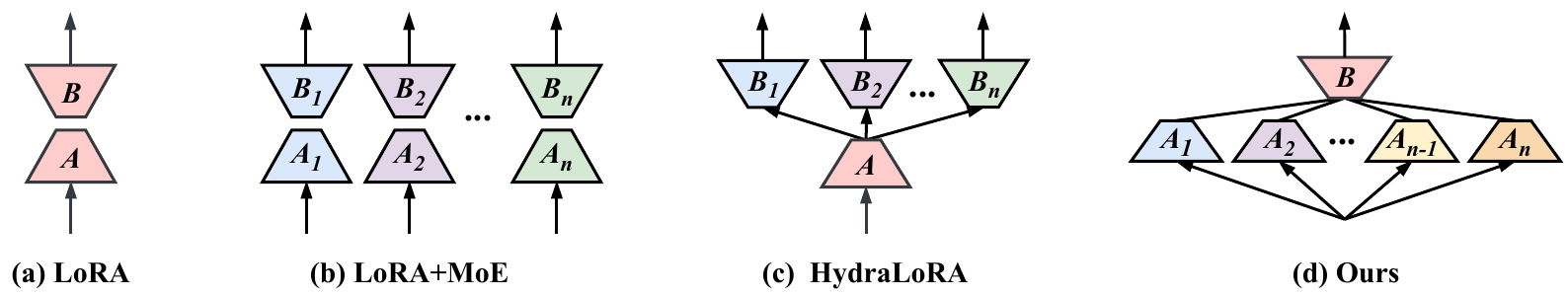}
  \caption{Architectures of LoRA variants and our proposed MASA.
  (a) \textbf{LoRA}: Each fine-tuned layer is augmented with a single pair of low‑rank adapters, one down‑projection $A$ and one up‑projection $B$.
  (b) \textbf{LoRA+MoE}: Model capacity is increased by instantiating $k$ independent adapter pairs $(A_i, B_i)$.
  (c) \textbf{HydraLoRA}: Several up‑projection heads $B_i$ share a common down‑projection $A$, forming a ``single‑$A$, multi‑$B$'' topology for parameter reuse.
  (d) \textbf{Ours}: We employ one shared up‑projection $B$ with multiple down‑projections $A_i$, which offers a balanced trade‑off between efficiency and representational capacity.
  }
  \label{fig:adapter_comparison}
\end{figure*}

Large Language Models (LLMs)~\cite{brown2020language,devlin2019bert,touvron2023llama,openai2023gpt} have demonstrated remarkable generalization capabilities across diverse tasks~\cite{raiaan2024review}, largely attributed to their ever-increasing scale.
Adapting these powerful models to specific downstream applications is commonly achieved through fine-tuning~\cite{han2024parameter}.
However, conventional Full Fine-Tuning (FFT) approaches update all model parameters, making them computationally expensive and memory-intensive, thereby posing significant challenges for resource-limited deployments.
As a solution, Parameter-Efficient Fine-Tuning (PEFT)~\cite{pfeiffer2020adapterfusion} has been proposed to keep the majority of the pre-trained model's parameters frozen and adjust only a small subset of new or existing parameters.
Among various PEFT methods such as Adapters~\cite{houlsby2019parameter} and Prefix-Tuning~\cite{li2021prefix}, Low-Rank Adaptation (LoRA)~\cite{hu2022lora} has gained particular prominence.

LoRA, a simple yet effective PEFT method, hypothesizes that the changes in weight during adaptation tend to be low-rank, which is implemented by injecting trainable low-rank matrices into transformer layers.
However, LoRA (shown in Fig.~\ref{fig:adapter_comparison}(a)) faces a fundamental low-rank bottleneck that consistently leads to performance gaps compared to full fine-tuning, particularly on complex reasoning tasks~\cite{biderman2024lora}. This limitation stems from the assumption of low-rank updating weights, which may not hold for tasks requiring substantial representation~\cite{mao2025survey}. When practitioners encounter insufficient performance, increasing the rank remains an alternative solution, but this directly increases the parameter and computational overhead, potentially compromising LoRA's efficiency advantages.

\begin{figure}[ht]
\vskip 0.2in
    \centering
    \includegraphics[width=\columnwidth]{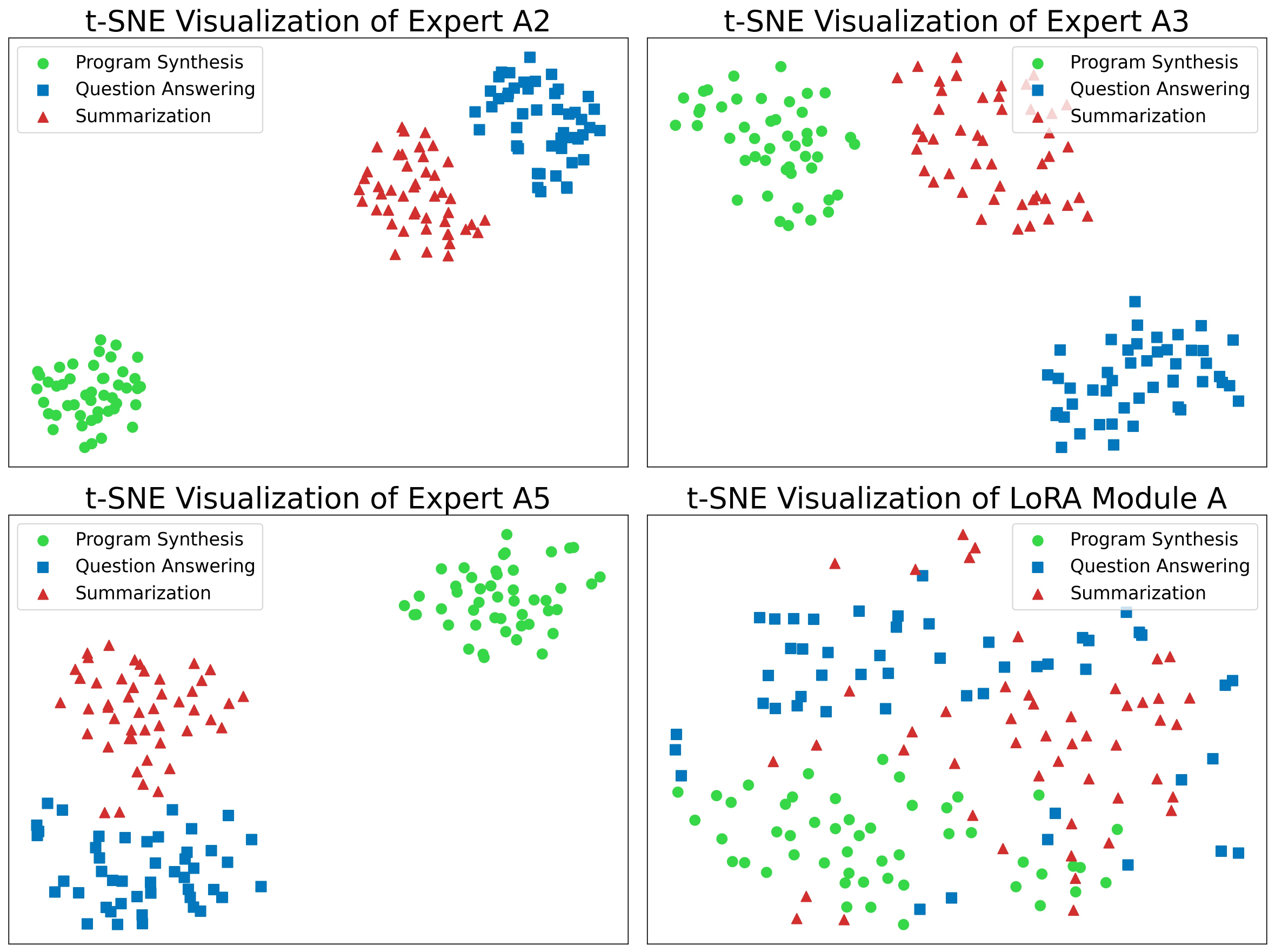} 
    \caption{
    The t-SNE visualization of task-specific features extracted from the V-projection layer of the 11$^{th}$ layer in the LLaMA3-8B model, comparing LoRA and three selected experts of our method after fine-tuning on OpenOrca.
        }
    \label{fig:tsne_comparison}
\end{figure}

To mitigate LoRA's limitations, the research community has explored various LoRA variants.
Initially, researchers pursued capacity expansion through the Mixture-of-Experts technique (\emph{e.g.}, LoRAMoE)~\cite{dou2024loramoe} (shown in Fig.~\ref{fig:adapter_comparison}(b)), which employs multiple parallel LoRA modules to decouple multi-task information. Although this enhances representational diversity by allowing different experts to specialize in distinct tasks, the independence of these experts limits the capture of shared knowledge and incurs substantial computation and parameter overhead. Subsequently, asymmetric architectures~\cite{tian2024hydralora,yang2025mtl} emerged as a more parameter-efficient solution, exemplified by HydraLoRA's ``one-$A$, multi-$B$'' architecture, as shown in Fig.~\ref{fig:adapter_comparison}(c). This design is motivated by the observation that the down-projection matrix ($A$) tends to learn shared, general features by adapting to diverse data, while the up-projection matrix ($B$) captures task-specific variations. This paradigm of allocating representational complexity to the matrix $B$ has become a prevailing design philosophy. However, such a ``single-$A$, multi-$B$'' strategy may introduce a potential representation bottleneck by relying on a single, low-rank $A$-matrix for all feature extraction. It appears that a single down-projection imposes a ceiling on the mutual information between the input and the update, thereby compressing every example into at most rank $r$ orthogonal directions (in Theorem~\ref{thm:bottleneck}).
Empirical results corroborate this phenomenon via t-SNE~\cite{maaten2008tsne}: As illustrated in Fig.~\ref{fig:tsne_comparison}, a single‑$A$ matrix maps examples from three disparate tasks into largely overlapping regions, whereas an ensemble of experts $A$ produces well‑separated clusters suitable for multi-task adaptation.

\begin{figure}[t]
\vskip 0.2in
    \centering
    \includegraphics[width=\columnwidth]{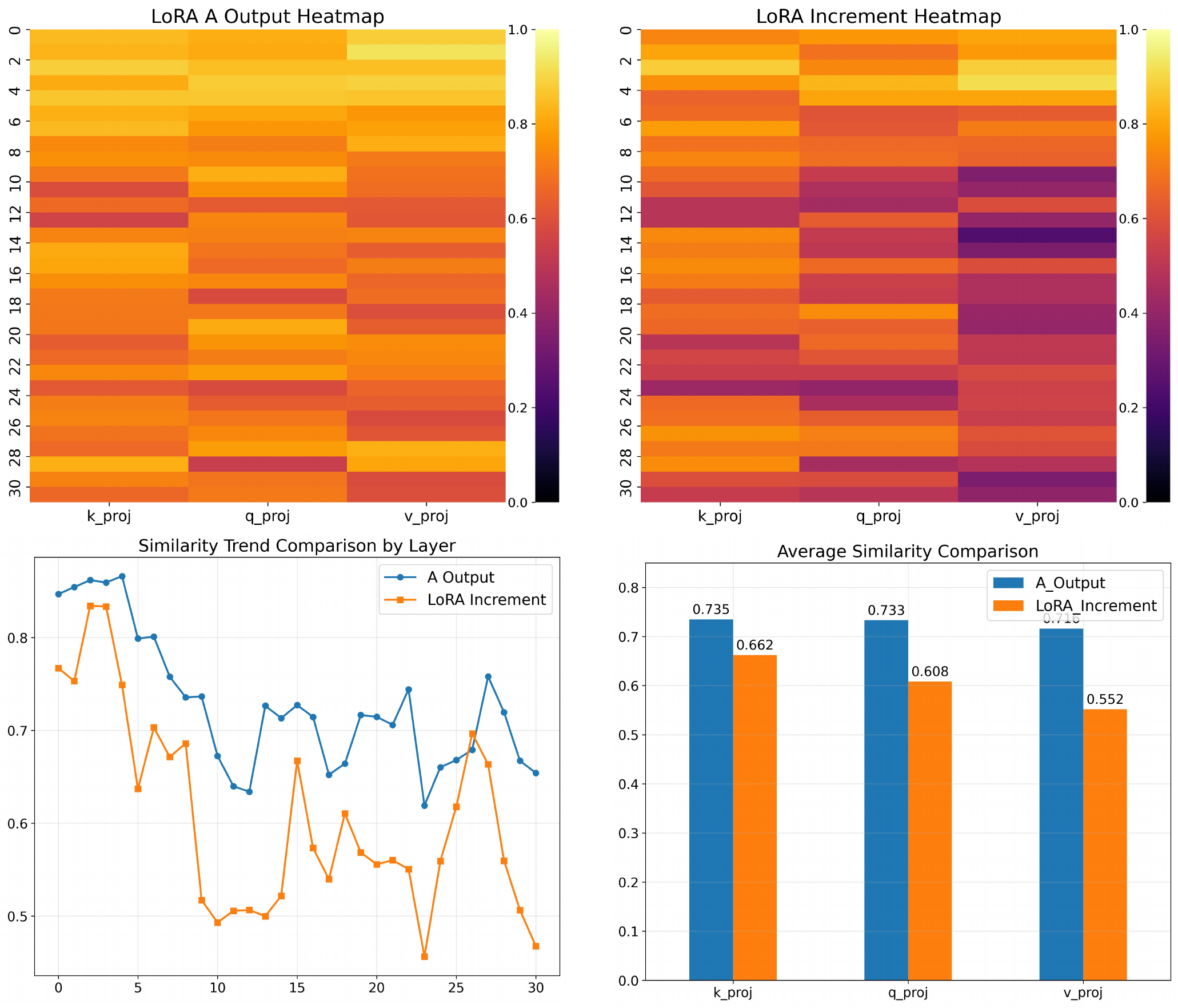} 
    \caption{Adjacent-layer similarity analysis of LoRA modules using the CKA method. (Top) Heatmaps of CKA similarity scores between adjacent layers for LoRA $A$-matrix outputs (Top left) and full LoRA increments (Top right). (Bottom left) Bar chart of average similarity scores by module type for $A$-matrix outputs versus LoRA increments. (Bottom right) Line plots of layer-wise similarity trends between consecutive layers throughout the network depth.}
    \label{fig:layerwise-similarity}
\end{figure}

The above analysis motivates us to overcome LoRA's representational bottleneck, requiring multiple down-projection matrices ($A$-experts) to capture diverse shared features, as shown in Fig.~\ref{fig:adapter_comparison} (d).
Meanwhile, the up-projection $B$ is used to integrate this knowledge for multi-domain and multi-task adaptations.
To maintain architectural simplicity and computational efficiency, we employ only one up-projection matrix, which is demonstrated to be sufficient for effective adaptation while significantly reducing parameter overhead.
Motivated by the observations of feature similarity across adjacent transformer layers~\cite{xiao2019sharing} and further corroborated by our adjacent-layer CKA~\cite{kornblith2019similarity}, we find that the outputs of $A$-experts are highly redundant across layers, indicating that these experts can be shared across layer groups (Fig.~\ref{fig:layerwise-similarity}).
To this end, we consider designing a parameter-efficient sharing mechanism, which not only improves adaptation efficiency but also enables the model to learn more generalized features. 

Guided by these design principles, we concretize them into Multi‑$A$ Shared Adaptation (MASA)—a compact yet effective adapter which breaks the representational bottleneck in LoRA by enriching the feature adaptation rather than simply adding multiple projection heads. As depicted in Fig.~\ref{fig:MASA_Framework}, MASA employs a Multi-$A$ Expert (MAE) design with a cross-layer sharing mechanism. MAE enables the model to extract diverse domain and task-specific features through an ensemble of experts $A$.
Furthermore, an Asymmetric Cross-layer Sharing (ACS) is proposed to reduce expert redundancy across layers, thereby implementing efficient adaptation.
The main contributions are summarized as:
\begin{itemize}
\item We propose Multi-$A$ Shared Adaptation (MASA), a novel asymmetric PEFT architecture that employs a ``multi-$A$, single-$B$'' structure to address the representational bottleneck in conventional LoRA methods. 
\item We introduce an Asymmetric Cross-layer Sharing (ACS) strategy that enables substantial parameter efficiency by sharing the $A$-matrix ensemble across adjacent layers while preserving layer-specific matrix $B$.
\item Comprehensive experiments on multiple benchmarks demonstrate that the proposed MASA achieves superior performance with parameter efficiency, compared to existing PEFT methods.
\end{itemize}

\section{Related Work}

\subsection{LoRA Architecture and Its Variants}
Parameter-Efficient Fine-Tuning (PEFT) has become essential for adapting Large Language Models (LLMs) to downstream tasks. Low-Rank Adaptation (LoRA)~\cite{hu2022lora} only adds and updates one trainable pair of low-rank matrices $A$ and $B$ in each layer during fine-tuning, while keeping the existing weights frozen. It has been one of the most effective solutions in PEFT. However, the monolithic nature of a single LoRA module, with its simple one-to-one connection, limits its capacity to capture the diverse features required for complex tasks.
This limitation spurred the development of Multi-LoRA architectures presented in Fig.~\ref{fig:adapter_comparison}. 
An early attempt, LoRAMoE~\cite{dou2024loramoe}, employs a Mixture-of-Experts framework with multiple LoRA modules. Although this enhances representational diversity, the independence of its experts impedes the learning of shared, domain-general knowledge.
Alternatively, subsequent research shifted towards asymmetric structures that share certain components.
Works like HydraLoRA~\cite{tian2024hydralora} and MTL-LoRA~\cite{yang2025mtl} adopt a ``one-$A$, multi-$B$'' paradigm, where a shared matrix $A$ captures common features and multiple matrices $B$ learn task-specific projections.
Even more flexible structures like CoLA~\cite{zhou2025cola}, which allow for many-to-many relationships, still predominantly adhere to a ``few-$A$, many-$B$'' philosophy. 
This places the primary burden of feature extraction on a smaller number of $A$-matrices, which can limit the model's ability to capture the diverse and nuanced signals required for complex tasks. Our work explores an alternative approach by shifting the architecture to focus on the feature extraction stage. We employ a ``multi-$A$, single-$B$'' structure, utilizing an ensemble of $A$-matrices to enrich the feature representations. 

\subsection{Cross-Layer Parameter Sharing Strategies}
Parameter sharing across layers has emerged as a fundamental strategy for reducing memory footprint and computational overhead while maintaining performance. ALBERT~\cite{lan2019albert} achieved up to 90\% parameter reduction in BERT by reusing attention and feed-forward weights across layers, while the Universal Transformer~\cite{dehghani2018universal} formalized depth as recurrent time steps with shared parameters. Similar successes in computer vision (LaViT~\cite{zhang2024lvit}) and inference optimization (KVSharer~\cite{yang2024kvsharer}) demonstrate reductions of up to 4× in parameters and 30\% in memory usage, respectively.
Building upon these insights, the LoRA community has explored parameter sharing strategies tailored to low-rank adaptation. As models scale, the cumulative parameter count across layers has motivated sharing investigations. Recent approaches include VeRA’s~\cite{kopiczko2023vera} sharing of random matrices, VB-LoRA’s~\cite{li2024vb} vector-bank reconstruction, and BSLoRA’s~\cite{zhoubslora} granular slicing schemes.
However, these methods face a fundamental limitation: their strategies assume symmetric LoRA structures, treating the $A$ and $B$ matrices equivalently. This prevents them from exploiting the functional asymmetry between feature extraction and projection identified. 
Guided by empirical analysis of inter-layer LoRA similarity, our method shares the down-projection matrices ($A$) across layer groups while preserving layer-specific up-projection matrices ($B$).  

\begin{figure}[t]
\vskip 0.2in
    \centering
    \includegraphics[width=1\columnwidth]{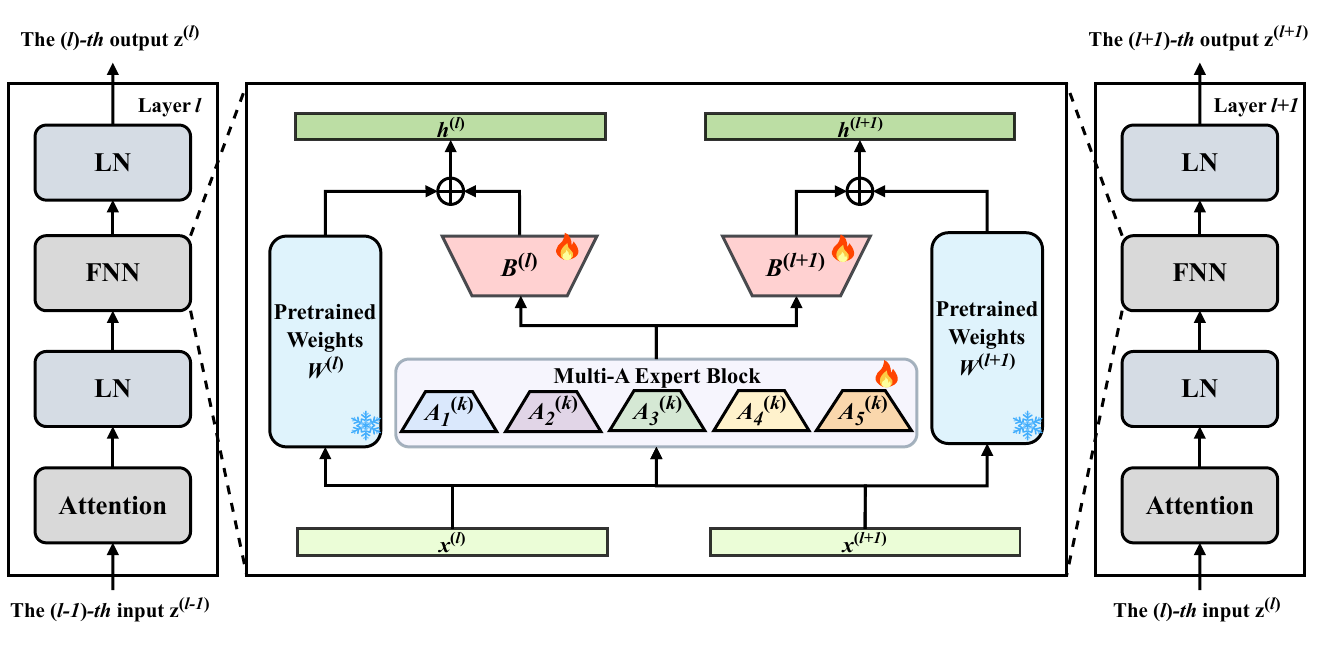}  
    \caption{
        An overview of our proposed MASA architecture. Each transformer layer is augmented with a shared set of $A$ modules and layer-specific $B$ modules. The shared $A$ modules are managed by a Multi-$A$ Expert (MAE) block, enabling inter-layer sharing for improved parameter efficiency via an Asymmetric Cross-layer Sharing (ACS). Pretrained weights ($W^{(i)}$) are frozen while only  $\{A_{i}^{(k)}\}$ ($k=\lfloor l/S \rfloor$) and $B^{(l)}$ are updated during fine-tuning, where $S$ is the group size.
    }
    \label{fig:MASA_Framework}
\end{figure}


\section{The Proposed Method}



\subsection{Preliminaries}

 Low-Rank Adaptation (LoRA)~\cite{hu2022lora} posits that the additional and updated weight $\Delta W$ is low-rank for a pre-trained weight matrix $W \in \mathbb{R}^{d_{out} \times d_{in}}$ during fine-tuning. Specifically, 
$\Delta W$ is decomposed into the low-rank matrices, $A \in \mathbb{R}^{r \times d_{in}}$ and $B \in \mathbb{R}^{d_{out} \times r}$, where the rank $r \ll \min(d_{in}, d_{out})$. During adaptation, $W$ remains frozen, and only $A$ and $B$ are trained. The modified forward pass for a hidden state $x$ is given by:
\begin{equation}
    h = W x + \Delta W x = W x + B A x.
\end{equation}
For initialization, the matrix $A$ is typically initialized with the Kaiming Uniform~\cite{he2015delving}, while $B$ is initialized to zero, ensuring that $\Delta W = BA$ is zero at the beginning of training.
Functionally, the LoRA module can be conceptualized as a two-stage process: feature compression and decompression.
The matrix $A$ acts as a projector or compressor, mapping the high-dimensional input features into a low-rank, compact space.
Subsequently, the matrix $B$ acts as a restorer or decompressor, projecting these low-rank features back to the original output dimension space.
For stable training, the LoRA update is typically scaled by a constant $\frac{\alpha}{r}$, where $\alpha$ is a learning rate-like scalar.
Thus, the final forward pass is $h = W_0 x + \frac{\alpha}{r} B A x$.





\subsection{A Rigorous Bottleneck of Single-Extractor LoRA}
\label{sec:bottleneck}

LoRA updates a frozen linear map via $\Delta W = BA$, where $A\in\mathbb{R}^{r\times d_{\text{in}}}$ acts as an
\emph{extractor} producing $\mathbf{u}=A\mathbf{x}\in\mathbb{R}^r$, and $B$ (or any deterministic $g$) only remixes
these $r$ features. Hence the adaptation branch is constrained by an $r$-dimensional bottleneck shaped by $A$.

\textbf{Finite precision / noisy representation.}
For continuous variables, mutual information under deterministic mappings may be ill-defined or infinite.
We therefore adopt a standard finite-precision/noisy-representation view:
\begin{equation}
\tilde{\mathbf{u}}=\mathbf{u}+\boldsymbol{\varepsilon},\qquad
\boldsymbol{\varepsilon}\sim\mathcal{N}(\mathbf{0},\sigma^2 I_r),\ \boldsymbol{\varepsilon}\perp \mathbf{x}.
\label{eq:noisy_u}
\end{equation}

\begin{theorem}[Information Ceiling (finite precision)]
\label{thm:bottleneck}
Let $\mathbf{x}\in\mathbb{R}^{d_{\text{in}}}$ have covariance $\Sigma_x\succeq 0$ and let $\mathbf{u}=A\mathbf{x}$
with $A\in\mathbb{R}^{r\times d_{\text{in}}}$. Define $\tilde{\mathbf{u}}$ as in~\eqref{eq:noisy_u}.
Then for any deterministic post-mapping $g$ (including $g(\tilde{\mathbf{u}})=B\tilde{\mathbf{u}}$ and multiple $B$-heads),
\begin{equation}
I(\mathbf{x};g(\tilde{\mathbf{u}}))
\;\le\;
\tfrac12\log\det\!\Bigl(I_r + \tfrac{1}{\sigma^2}A\Sigma_xA^{\top}\Bigr),
\quad
\mathrm{rank}(\mathbf{u})\le r .
\label{eq:bottleneck-short}
\end{equation}
\end{theorem}
\begin{proof}
Since $g$ is deterministic, $\mathbf{x}\to\tilde{\mathbf{u}}\to g(\tilde{\mathbf{u}})$ is a Markov chain, hence
$I(\mathbf{x};g(\tilde{\mathbf{u}}))\le I(\mathbf{x};\tilde{\mathbf{u}})$ by data processing.
Moreover, $\tilde{\mathbf{u}}=A\mathbf{x}+\boldsymbol{\varepsilon}$ with $\boldsymbol{\varepsilon}\perp\mathbf{x}$ gives
$I(\mathbf{x};\tilde{\mathbf{u}})=h(\tilde{\mathbf{u}})-h(\boldsymbol{\varepsilon})$.
The covariance is $\Sigma_{\tilde{u}}=A\Sigma_xA^\top+\sigma^2 I_r$.
By Gaussian maximum entropy under a covariance constraint,
$h(\tilde{\mathbf{u}})\le \frac12\log((2\pi e)^r\det(\Sigma_{\tilde{u}}))$ and
$h(\boldsymbol{\varepsilon})=\frac12\log((2\pi e)^r\det(\sigma^2 I_r))$,
yielding~\eqref{eq:bottleneck-short}. Finally, $\mathrm{rank}(\mathbf{u})\le \mathrm{rank}(A)\le r$.
\end{proof}

\subsection{The Whole MASA Framework}
\label{ssec:framework}

Based on Theorem~\ref{thm:bottleneck}, we find that the single matrix $A$ in standard LoRA and its asymmetric variants (\textit{e.g.}, HydraLoRA) may become a representational bottleneck, limiting the model's capacity to extract diverse and nuanced features. Moreover, the across-layer similarity on the feature extractor $A$ often incurs redundant computation.
%
%
To balance performance and efficiency, we design the Multi-$A$ Expert block with our asymmetric sharing strategy to construct \textbf{M}ulti-\textbf{$A$} \textbf{S}hared \textbf{A}daptation (MASA) framework. The overall architecture is illustrated in Fig.~\ref{fig:MASA_Framework}, which consists of two key modules, Multi-$A$ Expert Block (MAE) and Asymmetric Cross-layer Sharing (ACS). The shared matrices $A_i$ are managed by MAE, enabling inter-layer sharing to improve parameter efficiency via ACS. In detail, MAE and ACS are described in Section~\ref{MAE} and ~\ref{ACS}.

For a given layer $l$ in a MASA-adapted model, the modified forward pass is defined by its independent restoration matrix $B^{(l)}$ and the shared group of $A$-matrices $\{A_{i}^{(k)}\}_{i=1}^{N}$. The complete mathematical formulation is as follows:
\begin{equation} 
\small
    h^{(l)} = W_0^{(l)} x + \frac{\alpha}{r} B^{(l)} \left( \sum_{i=1}^{N} A_{i}^{(k)} \right) x, \quad \text{where } k=\lfloor l/S \rfloor.
\end{equation}
Here, $k$ is the index of the shared $A$-matrix group, determined by the layer $l$ and the size of sharing group $S$.


\subsection{The Multi-$A$ Expert Block}
\label{MAE}
As discussed above, our strategy is to reallocate the extra capacity to the feature extraction stage.
The Multi-$A$ Expert (MAE) block replaces the single matrix $A$ with an ensemble of $N$ expert matrices, $\{A_1, A_2, \dots, A_N\}$, where each $A_i \in \mathbb{R}^{r \times d_{in}}$. These experts operate in parallel, enabling the model to capture more diverse and fine-grained features from the input. We hypothesize that each expert matrix learns to specialize in distinct semantic subspaces. This functional specialization is empirically supported by t-SNE~\cite{maaten2008tsne} visualizations in Fig.~\ref{fig:tsne_comparison}, which illustrate that different $A_i$ matrices map the same inputs to separable feature clusters.
The outputs of these experts are aggregated and then projected back to the original dimension by a single restoration matrix $B \in \mathbb{R}^{d_{out} \times r}$. We aggregate expert features via summation, a simple yet effective approach that maintains parameter efficiency and training stability~\cite{xu2025smoothness}, with detailed rationale provided in Appendix~\ref{app:router-vs-sum}.
\begin{equation} 
    \Delta W_\text{MAE} = B \left( \sum_{i=1}^{N} A_i  \right)
\end{equation}
This ``multi-$A$, single-$B$'' design explicitly reallocates model capacity towards the crucial feature extraction stage, hypothesizing that a richer set of initial feature representations is more beneficial for complex tasks.


\subsection{Asymmetric Cross-layer Sharing}
\label{ACS}

To mitigate the potential parameter increase from using multiple $A$-matrices and to reduce their redundancy across transformer layers, we introduce an asymmetric cross-layer sharing mechanism for efficient adaptation. Unlike other sharing methods (VB-LoRA~\cite{li2024vb}, BSLoRA~\cite{zhoubslora}) that typically share the symmetric LoRA module (both $A$ and $B$), our approach introduces an asymmetric sharing strategy: we only share the ensemble of $A$-matrices across adjacent layers, while allowing each layer to retain its own independent, trainable $B$-matrix.


To substantiate our design, we conducted an empirical analysis of inter-layer activation similarity using Centered Kernel Alignment (CKA)~\cite{kornblith2019similarity}, a robust method for comparing neural network representations.
As presented in Fig.~\ref{fig:layerwise-similarity}, we applied CKA to measure the similarity between activation values from adjacent layers of a fine-tuned model, specifically comparing the $A$-matrix outputs ($Ax$) against the final LoRA increment activations ($BAx$). The analysis reveals a distinct pattern: The $A$-matrix outputs maintain a significantly higher similarity across layers than the full increments. This observation is quantified by the aggregated plots, which confirm a consistently higher average similarity score for the $A$-matrix outputs across all attention projections ($q, k, v$). This evidence indicates that 1) $A$-matrices learn to produce generalizable cross-layer representations, making them ideal candidates for sharing; and 2) $B$-matrices, conversely, perform more layer-specific transformations, which are critical for preserving the unique functionality of each layer.

Based on this empirical evidence, 
we group consecutive layers into blocks of size $S$. Within each block $k$, all layers share the same ensemble of $A$-matrices, $\{A_{1}^{(k)}, \dots, A_{N}^{(k)}\}$. However, each layer $l$ retains its own independent, trainable $B$-matrix, denoted as $B^{(l)}$. This approach leverages the generalizability of $A$-matrices for substantial parameter savings, while the layer-specific $B$-matrices preserve the model's capacity for nuanced, per-layer adaptation.

\subsection{Discussion on the benefits of Multi-$A$ Shared Extractors}
\label{sec:why_multiA}

Consider a multi-branch extractor $\{A_i\}_{i=1}^N$ with summation aggregation:
\begin{equation}
\sum_{i=1}^{N} A_i\mathbf{x} = A_{\text{sum}}\mathbf{x},
\qquad
A_{\text{sum}}\triangleq \sum_{i=1}^{N}A_i .
\label{eq:Asum}
\end{equation}
Thus, under summation, multi-$A$ is functionally equivalent to a single extractor $A_{\text{sum}}$ and does not
guarantee a larger ceiling in~\eqref{eq:bottleneck-short}. Its benefit is \emph{trainability}:
splitting $A_{\text{sum}}$ into multiple branches changes regularization and optimization dynamics and promotes
specialization under heterogeneous adaptation signals.

\textbf{Weaker effective $\ell_2$ regularization on the realized extractor.}
With per-branch weight decay $\lambda\sum_{i=1}^{N}\|A_i\|_F^2$, the induced penalty on $A_{\text{sum}}$ is weaker:
for any fixed $A_{\text{sum}}$, convexity implies
\begin{equation}
\sum_{i=1}^{N}\|A_i\|_F^2 \;\ge\; \frac{1}{N}\Big\|\sum_{i=1}^{N}A_i\Big\|_F^2
\;=\;\frac{1}{N}\|A_{\text{sum}}\|_F^2,
\label{eq:wd_short}
\end{equation}
with equality at $A_i=A_{\text{sum}}/N$. Hence multi-$A$ 
induces a weaker penalty on $A_{\text{sum}}$ in the optimal decomposition sense, allowing stronger extractor adaptation under the same rank-$r$ bottleneck.

\textbf{Reduced destructive gradient interference in an expanded parameter space.}
In multi-task/domain adaptation, gradients w.r.t. a shared extractor can conflict.
For a single $A$, the update aggregates task gradients $g=\sum_t g_t$, whose magnitude is diminished by negative
cross terms:
$\|g\|^2=\sum_t\|g_t\|^2+2\sum_{s<t}\langle g_s,g_t\rangle$.
With multiple branches, the over-parameterized decomposition provides additional degrees of freedom; under stochastic optimization, branches can follow different trajectories, which may alleviate cancellation in the effective update of the realized extractor.


\paragraph{The benefits of symmetry breaking.}
With summation aggregation, each layer uses
$A_{\text{sum}}^{(k)}=\sum_{i=1}^N A_i^{(k)}$ (Eq.~\ref{eq:Asum}), hence the objective is permutation-symmetric over experts.
In practice, random initialization and stochastic optimization break this symmetry, and experts are not forced to become
identical. For a layer $\ell$ in group $k$, $\Delta y_\ell = B_\ell A_{\text{sum}}^{(k)} x_\ell$ gives
\begin{equation}
\nabla_{A_i^{(k)}}\mathcal{L}
=\sum_{\ell\in\mathcal{G}(k)} \big(B_\ell^\top g_\ell\big)\,x_\ell^\top,
\qquad
g_\ell \triangleq \nabla_{\Delta y_\ell}\mathcal{L},
\label{eq:grad_acs_short}
\end{equation}
which is identical for all $i$ within the same group under summation.
Nevertheless, the over-parameterized decomposition together with symmetry breaking can sustain expert diversity in practice.
We empirically observe expert specialization (Fig.~\ref{fig:tsne_comparison}), which may help mitigate gradient interference
under heterogeneous adaptation signals.

\textbf{Low parameter complexity of MASA.} For a model with $L$ layers, the number of trainable parameters in MASA is approximately $\frac{L}{S} \times N \times d_{in} \times r + L \times d_{out} \times r$. The sharing mechanism significantly reduces the cost associated with the $A$-matrices, which would otherwise be the largest contributor due to multiple experts ($N$). This makes MASA's parameter count comparable to or even lower than standard LoRA ($L \times (d_{in} + d_{out}) \times r$), especially when $S$ is large. For example, on LLaMA3-8B, MASA in the $3A1B$ and $S=2$ configuration reduces parameters by $\sim$ 36.3\% compared to HydraLoRA in the corresponding $1A3B$ configuration.

\section{Experiments}


\begin{table*}[t]
\caption{Comparison of zero-shot performance (\%) of our proposed MASA method against other PEFT techniques on LLaMA3-8B. Performance is reported as accuracy on the MMLU benchmark and three single-domain tasks. The $A$ and $B$ specify the number of $A$ and $B$ matrices in the respective LoRA architectures. The best accuracy in each column is highlighted in \textbf{bold}. The results based on LLaMA3.2-3B are in Appendix~\ref{app:additional_results}.}
\centering
\setlength{\tabcolsep}{4pt} 
\begin{tabular}{lcccccccccccc}
\toprule
\multirow{2}{*}{\textbf{Model}} &
\multicolumn{5}{c}{\textbf{MMLU}} & 
\multicolumn{3}{c}{\textbf{Single Domain}} &
\multirow{2}{*}{\textbf{All Avg.}} & 
\multirow{2}{*}{\textbf{Params.} (\%)} & 
\multirow{2}{*}{\textbf{$A$}} & 
\multirow{2}{*}{\textbf{$B$}} \\
\cmidrule(lr){2-6}\cmidrule(lr){7-9}
& \textbf{Hum.} & \textbf{STEM} & \textbf{Social} & \textbf{Other} & \textbf{Avg.} & \textbf{Law} & \textbf{GSM8K} & \textbf{Finance} & & & & \\
\midrule
LLaMA3-8B      & 24.29 & 21.82 & 21.71 & 24.04 & 23.12 & 24.62    & 24.87    & 26.42    & 24.76 & --    & -- & -- \\
Prompt Tuning  & 25.08 & 23.88 & 24.37 & 23.27 & 24.26 & 24.90    & 25.40    & 24.53    & 24.77 & 0.0004 & -- & -- \\
P-Tuning      & 24.59 & 23.91 & 26.10 & 24.94 & 24.85 & 25.18    & 25.47    & 22.60    & 24.53 & 0.0280 & -- & -- \\
\midrule
LoRA\(_{r=8}\)       & 52.01 & 49.03 & 66.59 & 65.92 & 57.61 & \textbf{25.64} & 53.45 & 46.79 & 45.87 & 0.2605 & 1 & 1 \\
LoRA$_{r=16}$      & 52.41 & 50.46 & 68.05 & 66.59 & 58.54 & 25.07 & 56.62 & 46.79 & 46.76 & 0.5196 & 1 & 1 \\
DoRA\(_{r=16}\)      & 52.26 & 49.92 & 68.22 & 66.46 & 58.37 & 24.90 & 56.63 & 45.66 & 46.39 & 0.5366 & 1 & 1 \\
BSLoRA          & 52.14 & 48.72 & 67.24 & 65.98 & 57.74 & 24.56 & 54.97 & 41.51 & 44.70 & 0.2247 & 1 & 1 \\
VB-LoRA         & 52.24 & 49.64 & 66.72 & 66.01 & 57.88 & 24.84 & 57.69 & 42.64 & 45.76 & 0.2613 & 1 & 1 \\
CoLA           & 51.86 & 48.91 & 67.47 & 66.53 & 57.86 & 24.73 & 58.53 & 51.32 & 48.11 & 0.6551 & 2 & 3 \\
HydraLoRA      & 51.84 & 49.70 & 66.79 & 66.08 & 57.78 & \textbf{25.64} & 53.30 & 43.02 & 44.94 & 0.5785 & 1 & 3 \\
\midrule
\textbf{MASA}     & \textbf{53.28} & \textbf{50.65} & \textbf{69.81} & \textbf{68.23} & \textbf{59.62} & 25.41 & \textbf{58.98} & \textbf{52.45} & \textbf{49.12} & 0.5220 & 5 & 1 \\
\bottomrule
\end{tabular}

\label{tab:main_result}
\vskip -0.1in
\end{table*}

\subsection{Experimental Setups}


\textbf{Datasets and Benchmarks.}
All experiments are conducted on the pretrained LLaMA3-8B,  LLaMA3.1-8B and LLaMA3.2-3B models~\cite{grattafiori2024llama}. 
To evaluate multi-domain generalization, we fine-tune the model on the databricks-dolly-15k~\cite{conover2023free} dataset and assess its performance on the Massive Multitask Language Understanding (MMLU) benchmark~\cite{hendrycks2020measuring}. 
For single-domain specialization, we focus on three distinct professional fields. In Law, the model is fine-tuned on Lawyer-Instruct~\cite{Lawyer-Instruct} and US-Terms~\cite{li2023chatdoctor} and evaluated on the law-related tasks within MMLU. For Math, we use the training set of GSM8K~\cite{GSM8K} for fine-tuning and its test set for evaluation. Finally, in the Finance domain, we fine-tune on the training set of FinGPT-fineval~\cite{wang2023fingpt} and test on its corresponding evaluation set. For multi-task reasoning, we fine-tune the model on a subset of the OpenOrca~\cite{lian2023openorca} dataset and evaluate its performance on the Big-Bench Hard (BBH) benchmark~\cite{suzgun2022challenging}. 
Further details regarding these datasets are provided in Appendix~\ref{app:setting}.

\textbf{Implementation details.} 
Unless otherwise noted, we fix the LoRA rank to $r=8$, scaling factor to $\alpha=16$, number of $A$‑experts to $N=5$, and the size of sharing group to $S=2$. 
All models are fine‑tuned with AdamW. A full hyper‑parameter table is provided in Appendix~\ref{app:setting}.

\textbf{Baselines.}
To ensure a comprehensive evaluation, we compare the proposed MASA against representative methods spanning different PEFT paradigms: (1) Prompt and Representation Tuning: Methods that leave the backbone weights untouched and instead learn continuous prompts or prefix vectors: Prompt Tuning~\cite{lester2021power} and P\text{-}Tuning~\cite{liu2021gpt}. (2) Single–Adapter LoRA Family: All methods insert one lightweight adapter per modified module, differing only in how the low‑rank update is parameterised: LoRA~\cite{hu2022lora}, DoRA~\cite{liu2024dora}, VB‑LoRA~\cite{li2024vb}, and BS‑LoRA~\cite{zhoubslora}. (3) Multi–Adapter and Modular LoRA: Approaches that instantiate multiple LoRA‑style adapters and decide at run‑time how to combine them: LoRAMoE~\cite{dou2024loramoe}, HydraLoRA~\cite{tian2024hydralora}, MTL‑LoRA~\cite{yang2025mtl}, and CoLA~\cite{zhou2025cola}.

\begin{table*}[t]
\caption{Comparison of zero-shot performance using different PEFT methods on the BBH benchmark.}
\centering
\small
\setlength{\tabcolsep}{4pt}
\begin{tabular}{lccccccccc}
\toprule
\textbf{Model} & \textbf{Base} & \textbf{LoRA} & \textbf{LoRA$_{r=16}$} & \textbf{DoRA$_{r=16}$} & \textbf{MultiLoRA} & \textbf{MoELoRA} & \textbf{MTL-LoRA} & \textbf{HydraLoRA} & \textbf{MASA} \\
\midrule
LLaMA3-8B     & 29.96 & 41.80 & 42.00 & 41.96 & 40.85 & 41.04 & 42.23 & 40.94 & \textbf{42.36} \\
LLaMA3.1-8B   & 29.47 & 41.60 & 42.74 & 41.45 & 41.50 & 41.74 & 42.74 & 41.17 & \textbf{42.82} \\
\midrule
Num of $A$/$B$ & 0/0   & 1/1   & 1/1   & 1/1   & 2/2   & 2/2   & 1/3   & 1/3   & 5/1   \\
\midrule
Params. (\%)  & 0     & 0.2605 & 0.5196 & 0.5366 & 0.5196 & 0.5503 & 0.5327 & 0.5785 & 0.5220 \\
\bottomrule
\end{tabular}
\label{tab:bbh_result}
\vskip -0.1in
\end{table*}

\subsection{Main Results}
We compare our best-performing MASA configuration (5A1B with a sharing group size of $S=2$, denoted by MASA) against all baselines. Our main experimental results are presented in Tab.~\ref{tab:main_result} and Tab.~\ref{tab:bbh_result}. These findings, which span multi-domain generalization, single-domain specialization, and multi-task reasoning, consistently demonstrate that our proposed MASA framework achieves notable performance while maintaining parameter efficiency.






\textbf{Comparison results on MMLU.}
\label{ssubsec:mmlu_results}
As illustrated in Tab.~\ref{tab:main_result}, MASA demonstrates competitive advantages on the MMLU benchmark, which measures comprehensive model capabilities across diverse knowledge domains. Our method achieves an average accuracy of 59.62\%, outperforming all baseline methods, including LoRA variants and other PEFT approaches. Our MASA configuration achieves the highest accuracy among all compared methods while maintaining the number of additional trainable parameters at only 0.52\%, demonstrating parameter efficiency compared to competing methods like CoLA and HydraLoRA. This result provides compelling evidence that MASA's design effectively alleviates representational bottleneck to enhance generalization and reasoning capabilities across diverse domains.

\textbf{Comparison results on Single-Domain Specialization.}
The efficiency and effectiveness of MASA extend beyond general-purpose tasks and are consistently validated across highly specialized domains. As shown in Tab.~\ref{tab:main_result}, MASA exhibits superior performance in domain-specific tasks, achieving notable improvements in Finance, Math, and competitive performance in Law. These results demonstrate that MASA can effectively adapt the base model to achieve high performance in specialized areas while maintaining comparable parameter overhead. Averaging both multi-domain and single-domain tasks, the proposed MASA achieves the best performance of 49.12\% against all LoRA variants. This consistent performance across diverse specialized domains highlights MASA's effectiveness and broad applicability, 
making it particularly suitable for applications requiring domain-specific expertise.
\label{ssubsec:domain_results}

\textbf{Comparison results on Big-Bench Hard (BBH).}
To assess the model's complex reasoning capabilities, we evaluated MASA on the Big-Bench Hard (BBH) benchmark, which is specifically designed to challenge multi-step reasoning abilities. As detailed in Tab.~\ref{tab:bbh_result}, MASA consistently achieves leading performance across different baselines, reaching a top accuracy of 42.82\% on the LLaMA3.1-8B model while maintaining comparable parameter overhead of $\sim$ 0.52\%.
Specifically, MASA demonstrates a clear advantage over alternative multi-expert designs that follow the ``single-$A$, multi-$B$'' paradigm, such as HydraLoRA, and remains highly competitive with other methods like MTL-LoRA. The performance on BBH further validates that MASA's ``multi-$A$ architecture effectively alleviates the representational bottleneck, enhancing the model's capacity for complex problem-solving.

\textbf{Additional LLM backbones.}
To further verify robustness beyond LLaMA3, we evaluate MASA on two additional instruction-tuned LLMs:
\textbf{Qwen2.5-7B-Instruct} and \textbf{Mistral-7B} under the same MMLU protocol.
On Qwen2.5-7B-Instruct, MASA outperforms HydraLoRA and LoRA by +0.44pp and +0.20pp in average MMLU, respectively.
On Mistral-7B, MASA improves average MMLU by +0.81pp over HydraLoRA and +0.16pp over LoRA.
Results on LLaMA3.2-3B show consistent trends, and full breakdowns are provided in Appendix~\ref{app:additional_results}.

\subsection{Ablation Study and Analysis}


To understand why MASA is effective, we deconstruct its design through a series of targeted analytical and ablation experiments. These validate the two core principles: the ``multi-$A$, single-$B$'' structure and the Asymmetric Cross-layer Sharing (ACS) strategy.


\textbf{Ablation study on key modules.}
We first conduct an ablation study to quantify the individual contributions of MASA's key modules. As shown in Tab.~\ref{tab:ablation_MASA}, we compare three key configurations: (1) a standard \textbf{LoRA} baseline with 
rank set to 16; 
(2) \textbf{MAE}, which extends LoRA with the Multi-$A$ Expert Block (5$A$1$B$), while still keeping all layers independent.
(3) \textbf{MASA}, MAE with the Asymmetric Cross-layer Sharing.

The results reveal the synergy between the components. First, moving from standard LoRA to MAE yields a significant performance improvement, directly demonstrating the core value of the ``multi-$A$, single-$B$'' structure in enhancing the model's representational capacity. Subsequently, transitioning from MAE to MASA drastically reduces the number of trainable parameters while maintaining peak performance. 
This study confirms that our two core designs are indispensable: the Multi-$A$ structure is the performance engine, while Asymmetric Cross-layer Sharing is the efficiency safeguard.
\begin{table}[t]
\caption{Ablation study on the core modules of MASA on MMLU. The symbols \ding{51} and \ding{55} denote the presence and absence.}
\centering
\setlength{\tabcolsep}{4pt} 
\begin{tabular*}{\linewidth}{@{\extracolsep{\fill}}lcccc}
\toprule
\textbf{Model} & \textbf{MMLU Avg.} & \textbf{Params. (\%)} & \textbf{Multi-$A$} & \textbf{ACS} \\
\midrule
LoRA \(_{r=16}\) & 58.54 & 0.5196 & \ding{55} & \ding{55} \\
MAE              & 59.40 & 0.8280 & \ding{51} & \ding{55} \\
\textbf{MASA}    & \textbf{59.62} & \textbf{0.5220} & \ding{51} & \ding{51} \\
\bottomrule
\end{tabular*}
\label{tab:ablation_MASA}
\vskip -0.1in
\end{table}

\begin{table}[t]
\caption{Impact of varying the counts of $A$ matrices (\emph{left}, $B{=}1$) and $B$ projections (\emph{right}, $A{=}5$) on MMLU accuracy and parameter overhead. Experiments show that accuracy peaks at $A{=}5$ and then declines; increasing $B$ beyond one reduces accuracy while almost doubling the extra parameters, confirming the efficiency of the ``5$A$–1$B$'' design.}
\centering
\small
\begin{tabular*}{\linewidth}{@{\extracolsep{\fill} }@{}ccc@{\hspace{1.2em}}ccc@{}}
\toprule
\multicolumn{3}{c}{\textbf{Varying }$A$\textbf{ ($B=1$)}} &
\multicolumn{3}{c}{\textbf{Varying $B$ ($A=5$)}}\\
\cmidrule(lr){1-3} \cmidrule(lr){4-6}
$\!A$ & Avg. & Params. (\%)  & $\!B$ & Avg. & Params (\%) \\ \midrule
3 & 58.84 & 0.37   & 1 & \textbf{59.62} & \textbf{0.52} \\
4 & 59.46 & 0.45   & 2 & 59.36 & 0.73 \\
5 & \textbf{59.62} & \textbf{0.52} & 3 & 59.02 & 0.95 \\
6 & 59.13 & 0.60   & -- & -- & -- \\ \bottomrule
\end{tabular*}

\label{tab:vary_AB}
\vskip -0.1in
\end{table}



\textbf{Analysis of optimal $A$/$B$ configuration.}
We systematically explored the impact of different combinations of $A$ and $B$ matrices on the adaptation performance. As summarized in Tab.~\ref{tab:vary_AB}, we present the model's accuracy on the MMLU benchmark as a function of various ($A$, $B$) configurations.
\emph{Scaling $A$ with $B{=}1$.}
Increasing $A$ from $3$ to $5$ improves MMLU performance to 59.62\% with the modest parameter growth, while further scaling to $A{=}6$ declines the performance, indicating the performance saturation. \emph{Scaling $B$ with $A{=}5$.}
In contrast, increasing $B$ from $1$ to $3$ nearly doubles the parameter overhead while slightly reducing accuracy. This suggests that multiple $B$ projections introduce redundancy and slightly hurt performance.
These results validate the ``multi‑$A$, single‑$B$'' principle: a small ensemble of $A$ matrices can enrich the representation space, whereas duplicating $B$ mostly inflates the model. 


\textbf{Detailed effect of ACS.}
We further validate the rationale and optimality of our ACS. As the heatmaps in the methodology section revealed, $A$ matrices exhibit higher inter-layer similarity than $B$ matrices, which inspired our design to share only the $A$ matrices. To directly prove the superiority of this decision in terms of experimental performance, we conducted a comparative experiment across three different sharing strategies, which are summarized in Tab.~\ref{tab:share_strate}.

The results confirm that our design is well-motivated. The ``Share $A$, not $B$'' strategy achieves a favorable balance between performance and efficiency. In contrast, ``Share $B$, not $A$'' leads to a significant performance drop, suggesting the importance of layer-specific $B$ matrices for model expressiveness. Despite sharing both $A$ and $B$ improving parameter efficiency, it degrades performance from 59.62\% to 59.51\%, indicating that excessive sharing of $B$ may hinder layer-specific transformations. Overall, Asymmetric Cross-layer Sharing on the $A$-matrices is the best selection for adaptation.


To optimize our asymmetric sharing strategy, we evaluate different sharing group sizes ($S=1,2,4,8$). As shown in Tab.~\ref{tab:group_size_ablation}, $S=2$ achieves the best trade-off, yielding the highest MMLU accuracy (59.62\%) with only a 0.522\% learnable parameter. Larger $S$ values (4, 8) enhance efficiency but significantly reduce performance. These results validate that sharing $A$-matrices across adjacent layers ($S = 2$) is the optimal selection for MASA.

\begin{table}[t]
\caption{Effect of our Asymmetric Sharing strategy comparing against other potential sharing schemes.}
\centering
\begin{tabular*}{\linewidth}{@{\extracolsep{\fill} }l c c }
\toprule
\textbf{Sharing Strategy} & \textbf{MMLU Avg.} & \textbf{Params. (\%)} \\
\midrule
\textbf{MASA} & \textbf{59.62} & \textbf{0.5220} \\
\addlinespace 
Share $B$, not $A$ & 59.19 & 0.7605\\
\addlinespace
Share Both $A$ and $B$ & 59.51 & 0.4541 \\

\bottomrule
\end{tabular*}

\label{tab:share_strate}
\vskip -0.1in
\end{table}

\begin{table}[t]
\caption{Ablation study of the sharing group size (S) in ACS. We evaluate MASA's performance (MMLU Avg.) and parameter count under different layer sharing configurations. The optimal trade-off is achieved when $S=2$, yielding the highest MMLU accuracy while significantly reducing trainable parameters.}
\centering
\small
\begin{tabular*}{\linewidth}{@{\extracolsep{\fill} }ccc}
\toprule
\textbf{Group Size} & \textbf{MMLU Avg.} & \textbf{Params.  (\%)} \\
\midrule
1 & 59.40 & 0.828 \\
\textbf{2} & \textbf{59.62} & \textbf{0.522} \\
4 & 59.40 & 0.368 \\
8 & 59.18 & 0.291 \\
\bottomrule
\end{tabular*}
\label{tab:group_size_ablation}
\vskip -0.1in
\end{table}

\section{Conclusion}
\label{sec:conclusion}

In this paper, we identify the single down-projection matrix ($A$) in conventional LoRA-based methods as a key representational bottleneck. To address this, we propose MASA (Multi-$A$ Shared Adaptation), which employs an ensemble of specialized $A$-matrices with Asymmetric Cross-layer Sharing for enhanced feature extraction while reducing the learnable parameters. Comprehensive experiments across multi-domain generalization (MMLU), single-domain specialization (Law, Math, Finance), and multi-task reasoning (BBH) demonstrate that MASA consistently outperforms strong baselines. Ablation studies confirm that the multi-$A$ structure drives performance gains while asymmetric sharing ensures efficiency. Our findings suggest that reallocating the capacity to feature extraction presents a valuable direction for PEFT design.

\bibliography{example_paper}
\bibliographystyle{icml2026}

\newpage

\appendix

\section{Rationale for Summation-Based Aggregation}
\label{app:router-vs-sum}

\paragraph{Routers remain the de-facto choice in modern MoE.}
Token-level routers, introduced by the Sparsely-Gated layer\cite{shazeer2017outrageously} and subsequently adopt in GShard \cite{lepikhin2020gshard} and Switch-Transformer\cite{fedus2022switch}, enable conditional computation (top-$k$ expert selection) and explicit expert specialization. These benefits explain their popularity in large-scale language and vision models.

\paragraph{Dense aggregation is a competitive alternative when parameter / latency budgets are strict.}
Recent evidence suggests that averaging (either uniform or softly-weighted) can achieve comparable quality to hard routing while simplifying optimization:

\begin{itemize}
    \item SmoothMoE \cite{xu2025smoothness} replaces top-$k$ gating with a soft mask. With identical expert backends, the dense variant stays within $0.1$ BLEU of the routed model on WMT En-De while reducing gradient variance and eliminating auxiliary losses.
    \item On ImageNet, V-MoE~\cite{riquelme2021scaling}, reports that a dense (all-expert) mixture narrows the accuracy gap with routed ViT-MoE to less than $0.3$pp, while avoiding load-balancing issues.
\end{itemize}



\paragraph{Design Objectives of MASA: Capacity Reallocation with Architectural Simplicity.}
The proposed architecture prioritizes \textbf{feature-side} parameter enrichment through multi-$A$ experts while implementing \textbf{asymmetric cross-layer sharing (ACS)} to achieve parameter efficiency. The uniform summation mechanism, expressed as \(\Delta W_\text{MAE} = B \left( \sum_{i=1}^{N} A_i  \right)\), provides three key advantages:

\begin{enumerate}
    \item \textbf{Preservation of rank constraint}: The approach maintains the single rank-$r$ update structure, thereby preserving the theoretical foundations of bottleneck analysis;
    
    \item \textbf{Computational efficiency}: The method eliminates the need for routing parameters and avoids token-level branching operations, ensuring compliance with the latency requirements inherent to Parameter-Efficient Fine-Tuning (PEFT) methodologies;
    
    \item \textbf{Empirical sufficiency}: 
    While sparse routing strategies combined with effective load balancing have been proven to facilitate expert differentiation, our visualization results reveal that even a simple summation strategy can achieve effective expert specialization. The experimental results presented in Figure~\ref{fig:tsne_comparison} provide clear evidence of task differentiation among our Expert A matrices, demonstrating that summation mechanism is sufficient for achieving functional specialization.
\end{enumerate}

Given these trade-offs, we adopt summation as the simplest and most budget-friendly choice, while acknowledging that router-based aggregation remains attractive when resources permit larger expert counts.

\section{Additional Experimental Results}
\label{app:additional_results}

\begin{table*}[t]
\caption{Comparison of zero-shot performance (\%) of our proposed MASA method against other PEFT techniques on the LLaMA3.2-3B base model~\cite{grattafiori2024llama}. Performance is reported as accuracy on the MMLU benchmark, broken down into four domain subsets (Humanities, STEM, Social Sciences, Other) and their average. Columns $A$ and $B$ denote the number of $A$ and $B$ matrices in the respective LoRA architectures. The best accuracy in each column is highlighted in \textbf{bold}. }
\centering
\begin{tabular*}{\textwidth}{@{\extracolsep{\fill}} lcccccccc}

    \toprule
    \multirow{2}{*}{\textbf{Model}} & \multicolumn{5}{c}{\textbf{MMLU}} & \multirow{2}{*}{\textbf{Params.} (\%)} & \multirow{2}{*}{\textbf{$A$}} & \multirow{2}{*}{\textbf{$B$}}\\
    \cmidrule(lr){2-6}
    & \textbf{Hum.} & \textbf{STEM} & \textbf{Social} & \textbf{Other} & \textbf{Avg.} \\
    \midrule
    LLaMA3.2-3B     & 24.36 & 21.15 & 21.90 & 23.98 & 23.02 & --      & -- & -- \\
    Prompt Tuning   & 25.36 & 24.52 & 24.50 & 23.95 & 25.36 & 0.0008  & -- & -- \\
    P-Tuning        & 25.08 & 25.06 & 21.81 & 24.46 & 24.44 & 0.0530  & -- & -- \\
    \midrule
    LoRA\(_{r=8}\)       & 26.14 & 24.26 & 25.58 & 28.81 & 26.19 & 0.3770  & 1  & 1  \\
    LoRA\(_{r=16}\)      & 29.73 & 28.01 & 30.87 & 36.76 & 31.15 & 0.7511  & 1  & 1  \\
    DoRA\(_{r=16}\)     & 28.25 & 26.26 & 28.53 & 33.38 & 29.00 & 0.7748  & 1  & 1  \\
    CoLA            & \textbf{36.28} & 32.25 & 40.88 & 46.44 & 38.63 & 0.9406  & 2  & 3  \\
    HydraLoRA       & 31.52 & 28.16 & 32.95 & 38.17 & 32.55 & 0.8267  & 1  & 3  \\
    \midrule
    MASA      & 35.96 & \textbf{32.70} & \textbf{42.31} & \textbf{46.93} & \textbf{39.05} & 0.8810  & 5  & 1  \\
    \bottomrule
\end{tabular*}
\label{tab:mmlu_3B}
\vskip -0.1in
\end{table*}
\subsection{Additional Backbone Results}

Table~\ref{tab:mmlu_3B} complements the results in Section~4 by benchmarking all Parameter-Efficient Fine-Tuning (PEFT) baselines on the smaller 3 B-parameter LLaMA3.2-3B~\cite{grattafiori2024llama} backbone. Consistent with the findings on LLaMA3-8B~\cite{grattafiori2024llama}, our method MASA achieves the highest zero-shot accuracy on MMLU~\cite{hendrycks2020measuring}, reaching an average of 39.05\,\%. This represents a 0.42pp gain over second best baseline, CoLA~\cite{zhou2025cola}, while requiring 6.3\,\% fewer trainable parameters. MASA also secures column-leading results in three of the four domain subsets (STEM, Social Sciences, and Other) with a configuration of \(A=5\) and \(B=1\). Relative to LoRA~\cite{hu2022lora} and HydraLoRA~\cite{tian2024hydralora}, MASA delivers gains of 7.9 and 6.5 percentage points, respectively. Preservation of this performance across both the 8B and 3B backbones substantiates the scalability and robustness of MASA under varying model capacities and parameter budgets.

We further evaluate MASA on Qwen2.5-7B-Instruct and Mistral-7B; the per-category MMLU breakdowns are reported in Table~\ref{tab:extra_backbones_mmlu}.
Due to baseline availability and compute budget, these additional-backbone runs compare MASA against the strongest applicable PEFT baselines (LoRA and HydraLoRA) under the same rank setting.

\begin{table*}[t]
\caption{
Additional backbone results on MMLU. We report average MMLU and three subject groups.
All methods use the same rank ($r=16$). $(\#A,\#B)$ indicates the numbers of down-projection matrices $A$ and up-projection matrices $B$.
}
\centering
\setlength{\tabcolsep}{4.6pt}
\begin{tabular}{lccccc}
\toprule
\multicolumn{6}{c}{\textbf{Qwen2.5-7B-Instruct} (MMLU)}\\
\cmidrule(lr){1-6}
Method & Avg & Law & Medical & Other & $(\#A,\#B)$ \\
\midrule
LoRA       & 71.71 & 55.30 & 73.79 & 74.10 & (1,1) \\
HydraLoRA  & 71.47             & 54.74             & 72.70             & 72.03             & (1,3) \\
\textbf{MASA} & \textbf{71.91}    & \textbf{55.53}    & \textbf{74.40}    & \textbf{74.24}    & (5,1) \\
\midrule
\multicolumn{6}{c}{\textbf{Mistral-7B} (MMLU)}\\
\cmidrule(lr){1-6}
Method & Avg & Law & Medical & Other & $(\#A,\#B)$ \\
\midrule
LoRA       & 58.83 & 47.93 & 64.03 & 59.90 & (1,1) \\
HydraLoRA  & 58.18             & 47.93 & 63.21             & 59.16             & (1,3) \\
\textbf{MASA} & \textbf{58.99}    & \textbf{48.33}    & \textbf{64.71}    & \textbf{59.96}    & (5,1) \\
\bottomrule
\end{tabular}

\label{tab:extra_backbones_mmlu}
\vskip -0.1in
\end{table*}

\subsection{Expert Matrix Specialization Analysis}
\begin{figure*}[t]
\vskip 0.2in
    \centering
    \includegraphics[width=\linewidth]{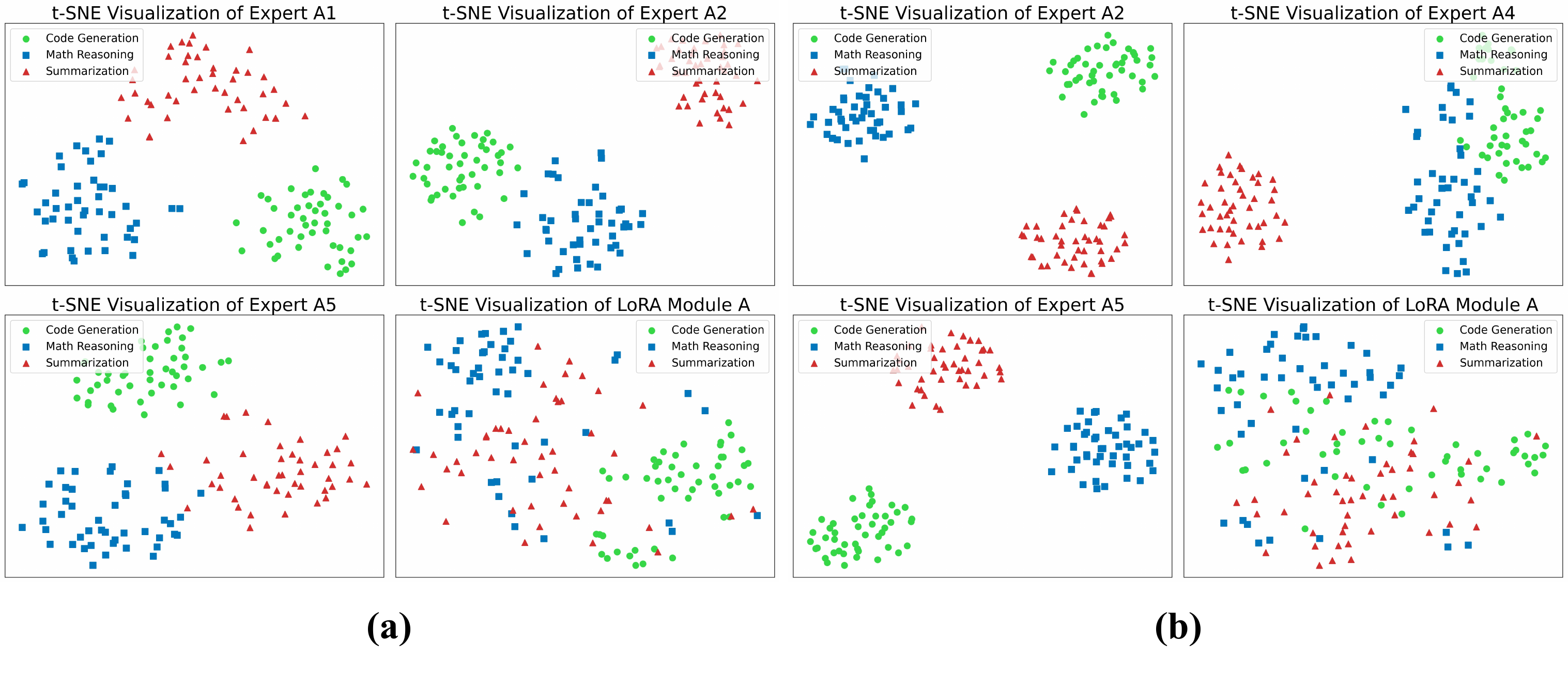} 
    \caption{
    The t-SNE visualization of task-specific features extracted from the Q-projection layer of the (a) 11$^{th}$ and (b) 14$^{th}$ layer in the LLaMA3-8B model, comparing LoRA and three selected experts of our method after fine-tuning on OpenOrca.
    }
    \label{fig:vis}
\end{figure*}

\textbf{t-SNE Visualization of Expert Matrices.}
Figure~\ref{fig:vis}(a) presents t-SNE visualizations of the Q-module A matrices at layer 11, specifically examining Expert A1, Expert A2, and Expert A5. Each subplot displays the feature distributions for three distinct task categories: Code Generation (represented by green circles), Math Reasoning (represented by blue squares), and Summarization (represented by red triangles). The visualizations reveal that each expert matrix exhibits clear functional specialization by mapping identical input representations to well-separated feature clusters in the embedding space. Notably, Expert A1 demonstrates strong cluster separation with Code Generation tasks forming a distinct cluster in the upper-right region, while Math Reasoning and Summarization tasks occupy separate regions in the lower portion of the feature space. Expert A2 shows a different clustering pattern, with Math Reasoning tasks concentrated in the center-left region and Code Generation tasks distributed in the upper-right area. Expert A4 exhibits yet another specialization pattern, with each task type forming compact, non-overlapping clusters in distinct regions of the feature space.


Figure~\ref{fig:vis}(b) presents the t-SNE visualizations for layer 14, focusing on Expert A2, Expert A4 and Expert A5. Consistent with the layer 11 analysis, the three task types maintain their color coding scheme throughout the visualization. Experts A2, A4, and A5 all demonstrate pronounced task-specific clustering characteristics, with each expert matrix capable of mapping different task types to distinct regions in the feature space, forming clearly separated clustering groups. In contrast, LoRA Module A exhibits significantly inferior clustering performance compared to these specialized expert matrices: while the LoRA module still achieves some degree of task separation, its clustering boundaries are notably more ambiguous, with greater overlap between different task types, and both cluster compactness and separability are less distinct than those observed in Experts A2, A4, and A5. This indicates that expert matrices in our proposed MASA possess superior functional specialization capabilities, enabling them to more effectively create unique representational spaces for different task types.

\textbf{Implications for Functional Specialization.}
These t-SNE visualizations provide empirical validation for our hypothesis that expert matrices learn to specialize in distinct semantic subspaces. The consistent emergence of task-specific clusters across different layers (11 and 14) and different expert matrices demonstrates that the functional specialization is not merely a surface-level phenomenon but represents a fundamental characteristic of the expert-based architecture.

The separability of feature clusters indicates that each expert matrix develops unique representational capabilities tailored to specific task domains. This specialization enables the model to efficiently allocate computational resources and maintain distinct processing pathways for different types of linguistic and cognitive tasks. Furthermore, the preservation of clustering patterns across network depths suggests that this specialization is maintained throughout the forward pass, contributing to the overall effectiveness of the mixture-of-experts approach.


\section{Experimental Setting}
\label{app:setting}

\subsection{Datasets and Benchmarks}
In our experiments, we train our method on 6 diverse datasets spanning multi-domain generalization, single-domain specialization, and multi-task reasoning. Table~\ref{tab:datasets} provides an overview of the datasets employed in this study. The detailed characteristics of each dataset are described as follows.
\begin{itemize}

\item \textbf{databricks-dolly-15k}: databricks-dolly-15k is an open-source dataset of instruction-following records containing more than 15,000 human-generated prompt/response pairs authored by over 5,000 databricks employees during March and April 2023. The dataset covers eight behavioral categories outlined in the InstructGPT paper, including brainstorming, classification, closed QA, generation, information extraction, open QA, and summarization. 

\item \textbf{Lawyer-Instruct}: Lawyer-Instruct is a conversational dataset primarily in English, reformatted from the original LawyerChat dataset. It contains legal dialogue scenarios reshaped into an instruction, input, and expected output format, making it ideal for supervised dialogue model training in the legal domain. The dataset focuses on legal question-answering scenarios and provides structured responses for legal inquiries and consultations.

\item \textbf{US-Terms}: US-Terms is one of the eight sub-tasks within LegalLAMA, a comprehensive benchmark suite designed to evaluate the extent of legal knowledge acquired by pre-trained language models during pre-training. LegalLAMA serves as a diverse probing benchmark designed to assess the legal knowledge that PLMs have acquired, with US-Terms specifically focusing on evaluating understanding of legal terminology and concepts within the United States legal system. The benchmark facilitates detailed analysis of legal-oriented language models' capabilities in legal knowledge comprehension.

\item \textbf{GSM8K}: GSM8K (Grade School Math 8K) is a dataset of 8,500 high-quality, linguistically diverse grade school math word problems created by human problem writers. The dataset is segmented into 7,500 training problems and 1,000 test problems that require 2 to 8 steps to solve, with solutions primarily involving elementary arithmetic operations. Problems require no concepts beyond the early algebra level and are designed such that a bright middle school student should be able to solve every problem. The dataset was specifically created to diagnose failures in multi-step mathematical reasoning capabilities of large language models, providing solutions in natural language format rather than pure mathematical expressions.

\item \textbf{FinGPT-fineval}: FinGPT-fineval is a Chinese multiple-choice questions dataset containing 1,060 training examples and 265 test examples, developed as part of the FinGPT benchmark for financial domain evaluation. The dataset consists of financial knowledge questions designed to assess large language models' understanding of financial concepts and reasoning capabilities in Chinese financial contexts. It serves as a component of the comprehensive FinGPT instruction tuning paradigm for open-source financial large language models.

\item \textbf{OpenOrca}: OpenOrca is a large-scale dataset containing approximately 1 million GPT-4 completions and 3.2 million GPT-3.5 completions, designed as an open reproduction of the dataset described in Microsoft Research's Orca paper. The dataset consists of augmented FLAN Collection data that aligns with the distributions outlined in the original Orca research, which focuses on progressive learning from complex explanation traces. OpenOrca enables training of smaller models that can achieve significant performance improvements on complex reasoning tasks through step-by-step explanation learning and instruction following.

\end{itemize}

The MMLU and BBH benchmarks employed in our experiments are described below.
\begin{itemize}
    \item \textbf{MMLU}:
Designed to probe broad world knowledge and problem-solving skills across 57 academic and professional subjects at varying difficulty levels.  

    \item \textbf{BBH}:
A challenging 23-task slice of BIG-Bench emphasizing deliberate multi-step reasoning (e.g., date understanding, logic grid puzzles, causal judgement).  
We follow the creators’ zero-shot protocol (no few-shot examples, no chain-of-thought disclosure) for strict comparability.
\end{itemize}

Some fine-tuning datasets (such as GSM8K and subsets of BBH) employ instruction formats that are not originally designed as multiple-choice questions, thus requiring the addition of answer options to accommodate classification task. We follow the approach used in COLA to meet the requirements.
\begin{table*}[t]
\caption{Overview of the datasets employed in this study.}
  \centering
  \small
  \setlength\tabcolsep{3pt}  
  \begin{tabularx}{\textwidth}
    {>{\raggedright\arraybackslash}p{2.7cm}  
     >{\raggedright\arraybackslash}p{1.3cm}   
     >{\raggedright\arraybackslash}p{4.5cm}   
     X X}                                     
    \toprule
    \textbf{Dataset}  & \textbf{Lang.} & \textbf{Source / Version} & \textbf{Task Types} \\
    \midrule
    databricks-dolly-15k &
      EN &
      Databricks v2 (Mar–Apr 2023) &
      Brain-storming, content generation, classification, summarization, information extraction, closed/open QA  \\
    \addlinespace[2pt]
    Lawyer-Instruct &
      EN &
      Public {\it Lawyer-Instruct} release (re-structured {\it LawyerChat}) &
      Multi-turn legal consultation dialogue  \\
    \addlinespace[2pt]
    US-Terms &
      EN &
      {\it LegalLAMA} v1.0 sub-task “US-Terms” &
      Legal terminology completion / knowledge recall \\
    \addlinespace[2pt]
    GSM8K &
      EN &
      Official train + test split &
      Multi-step grade-school math reasoning  \\
    \addlinespace[2pt]
    FinGPT-fineval &
      ZH$\rightarrow$EN &
      {\it FinGPT} fineval-v1 &
      Financial knowledge multiple-choice questions  \\
    \addlinespace[2pt]
    OpenOrca &
      EN &
      Snapshot 2024-06-15 &
      General instruction following (reasoning, dialogue, code, retrieval, \ldots) \\
    \bottomrule
  \end{tabularx}
  
  \label{tab:datasets}
  \vskip -0.1in
\end{table*}

\subsection{Baselines}
In the experiments, we compare our method against 11 representative PEFT baseline methods, covering various paradigms including prompt-based tuning, low-rank adaptation, and mixture-of-experts approaches. These baseline methods are described as follows:
\begin{itemize}
    \item \textbf{Prompt Tuning}: Prompt Tuning~\cite{lester2021power} is a parameter-efficient fine-tuning method that learns "soft prompts" - continuous prompt embeddings that are prepended to the input of frozen language models. Unlike discrete text prompts, soft prompts are learned through backpropagation and can incorporate signals from labeled examples. The method becomes more competitive with scale, matching full fine-tuning performance as models exceed billions of parameters while using significantly fewer trainable parameters.

    \item \textbf{P-Tuning}: P-Tuning~\cite{liu2021gpt} employs trainable continuous prompt embeddings to enhance the natural language understanding capabilities of GPT-style models. The method introduces learnable virtual tokens that are optimized during fine-tuning while keeping the pre-trained model parameters frozen. P-Tuning v2 extends this concept with deep prompt tuning, applying continuous prompts to every layer input of the transformer, significantly improving performance universally across scales and tasks.

    \item \textbf{LoRA}: Low-Rank Adaptation (LoRA)~\cite{hu2022lora} freezes pre-trained model weights and injects trainable rank decomposition matrices into each layer of the transformer~\cite{vaswani2017attention} architecture. The method hypothesizes that weight updates during adaptation have a low intrinsic rank, decomposing the update matrix into the product of two low-rank matrices. LoRA dramatically reduces trainable parameters while maintaining comparable performance to full fine-tuning.

    \item \textbf{DoRA}: Weight-Decomposed Low-Rank Adaptation (DoRA)~\cite{liu2024dora} decomposes pre-trained weights into magnitude and direction components for fine-tuning. It employs LoRA for directional updates while separately training the magnitude vector, effectively enhancing both learning capacity and training stability of LoRA. DoRA consistently outperforms LoRA across various downstream tasks while avoiding additional inference overhead.

    \item \textbf{VB-LoRA}: VB-LoRA~\cite{li2024vb} introduces a ``divide-and-share'' paradigm that breaks the barriers of low-rank decomposition across matrix dimensions, modules, and layers by sharing parameters globally via a vector bank. The method composes all the low-rank matrices of LoRA from a shared vector bank with a differentiable top-k admixture module. VB-LoRA achieves extreme parameter efficiency while maintaining comparable or better performance compared to state-of-the-art PEFT methods, using only 0.4\% of LoRA's stored parameters when fine-tuning large language models.

    \item \textbf{BSLoRA}: Bi-Share LoRA (BSLoRA)~\cite{zhoubslora} extends local LoRA with intra-LoRA and inter-LoRA parameter sharing to better capture both local and global information simultaneously. The method reduces trainable parameters through both intra-layer and inter-layer parameter sharing while maintaining or even enhancing model performance. BSLoRA incorporates three transformation methods to improve the compatibility and collaborative efficiency of shared parameters with varying shapes, achieving superior performance with only 44.59\% of the parameters of standard LoRA.

    \item \textbf{CoLA}: Collaborative Low-Rank Adaptation (CoLA)~\cite{zhou2025cola} introduces a flexible LoRA architecture that decouples the number of low-rank matrices A and B, enabling collaborative strategies for multi-task learning. CoLA employs task-agnostic and task-specific modules with three collaborative strategies: fully collaborative, random collaborative, and heuristic collaborative approaches, effectively mitigating task interference while maintaining parameter efficiency.

    \item \textbf{HydraLoRA}: HydraLoRA~\cite{tian2024hydralora} is an asymmetric fine-tuning architecture that effectively identifies and adapts to intrinsic data components, such as sub-domains or diverse tasks. It allocates distinct B matrices for task-specific features, while a shared A matrix integrates global information, enabling efficient parameter utilization and enhanced performance. The architecture employs a trainable MoE router to automatically segregate training samples and dynamically merge multiple B matrices during inference.

    \item \textbf{MultiLoRA}: MultiLoRA~\cite{wang2023multilora} addresses the limitation of LoRA's explicit low-rank constraint in complex multi-task scenarios by reducing the dominance of the top singular vectors. The method scales LoRA modules horizontally and changes parameter initialization to reduce parameter dependency, yielding more balanced unitary subspaces. MultiLoRA democratizes LoRA adaptation for better multi-task learning by enabling more effective knowledge sharing across diverse tasks.

    \item \textbf{LoRAMoE}: LoRAMoE~\cite{dou2024loramoe}  introduces several low-rank adapters (LoRA) and integrates them using a router network, functioning as a plugin version of Mixture of Experts (MoE). The method freezes the backbone model and forces a portion of LoRAs to focus on leveraging world knowledge to solve downstream tasks, effectively alleviating world knowledge forgetting during supervised fine-tuning. LoRAMoE addresses the challenge that large-scale increases in instruction data can damage the world knowledge previously stored in large language models.

    \item \textbf{MTL-LoRA}: Multi-Task Learning LoRA (MTL-LoRA)~\cite{yang2025mtl} enhances LoRA's multi-task learning capability by incorporating task-specific transformations in low-rank space along with adaptive exploration of multiple information sharing methods. The approach augments LoRA with additional task-adaptive parameters that differentiate task-specific information while capturing shared knowledge across various tasks within low-dimensional spaces, enabling effective joint adaptation to different target domains.
    \end{itemize}

\subsection{Hyperparameters}

We carefully tuned the hyperparameters for optimal model performance, with the complete configuration detailed in Table~\ref{tab:hyperparams}. The model was trained with a batch size of 8,and gradient accumulation steps of 8, yielding an effective batch size of 64. This configuration balances memory efficiency with stable gradient updates under GPU memory constraints.

The training process was conducted over 5 epochs with a validation split of 0.1, employing a step-based evaluation strategy to monitor model performance throughout training. We adopted a learning rate of 5e-5, paired with the AdamW optimizer and a cosine learning rate scheduler, a combination that has demonstrated superior performance in large language model fine-tuning tasks.

To ensure efficient processing of variable-length inputs, we set the maximum sequence length to 1024 tokens. For computational efficiency, we employed FP16 mixed precision training, which significantly reduces memory consumption while accelerating the training process. The entire training procedure was executed on 2 NVIDIA A6000 GPUs (48GB VRAM each), leveraging multi-GPU parallel computing capabilities.

These hyperparameter choices represent a careful balance between model performance, training stability, and computational resource constraints, as summarized in Table~\ref{tab:hyperparams}.

\begin{table}[ht]
\caption{Experimental hyperparameter settings.}
\centering  
\begin{tabular}{ll}  
\toprule
\textbf{Hyperparameter}                & \textbf{Setting} \\ \midrule
Batch Size                             & 8 \\
Train Epochs                           & 5.0 \\
Validation Size                        & 0.1 \\
Learning Rate                          & 5e-5 \\
Cutoff Length                          & 1024 \\
Gradient Accumulation Steps            & 8 \\
Scheduler Type                         & cosine \\
Precision                              & fp16 \\
Evaluation Strategy                    & steps \\
Optimizer                              & AdamW \\
GPU                                    & 2 A6000 (48G) GPUs \\ \bottomrule
\end{tabular}
\label{tab:hyperparams}
\vskip -0.1in
\end{table}
\label{subsec:hyperparameters}  

\end{document}